\documentclass[times,review,10pt]{elsarticle}
\usepackage{amsmath,amsfonts}
\usepackage{algorithmic}
\usepackage{algorithm}
\usepackage{array}
\usepackage[caption=false,font=normalsize,labelfont=sf,textfont=sf]{subfig}
\usepackage{textcomp}
\usepackage{stfloats}
\usepackage{url}
\usepackage{verbatim}  
\usepackage{graphicx}
\usepackage{cite}
\usepackage{color}
\usepackage{units}
\usepackage{amsthm,amssymb}
\usepackage{mathrsfs}
\usepackage{booktabs}
\usepackage{multirow}
\usepackage{pifont}
\usepackage{indentfirst}

\usepackage[markup=default,authormarkup=none,commentmarkup=footnote,commandnameprefix=always]{changes}
\definechangesauthor[name={revision}, color=teal]{revision}

\usepackage{hyperref}  
\hypersetup{hidelinks,
	colorlinks=true,
	allcolors=black,
	pdfstartview=Fit,
	breaklinks=true}

\newcommand{\xmark}{\ding{55}}%

\journal{Pattern Recognition}
\begin{document}
\begin{frontmatter}
\pdfstringdefDisableCommands{%
  \def\corref#1{*}  
}

\title{A Spatial Semantics and Continuity Perception Attention for Remote Sensing Water Body Change Detection}
\tnotetext[1]{Quanqing Ma and Jiaen Chen contribute equally to the paper.}
\author[mymainaddress]{Quanqing Ma}
\ead{m824454857@outlook.com}

\author[mymainaddress]{Jiaen Chen}
\ead{jiaenchen2024@outlook.com}

\author[mymainaddress]{Peng Wang}
\ead{17560664479@163.com}

\author[mymainaddress]{Yao Zheng\corref{cor}}
\ead{zy_inf@shzu.edu.cn}

\author[mymainaddress,hismainaddress]{Qingzhan Zhao}
\ead{zqz_inf@shzu.edu.cn}

\author[mymainaddress]{Yuchen Zheng\corref{cor}}
\ead{ouczyc@outlook.com}

\cortext[cor]{Corresponding authors.}

\address[mymainaddress]{College of Information Science and Technology, Shihezi University, Shihezi 832000, China }
\address[hismainaddress]{Production and Construction Corps Engineering Research Center for Spatial Information Technology, Shihezi 832000, China}

\begin{abstract}
Remote sensing Water Body Change Detection (WBCD) aims to detect water body surface changes from bi-temporal images of the same geographic area. Recently, the scarcity of high spatial resolution datasets for WBCD restricts its application in urban and rural regions, which require more accurate positioning. Meanwhile, previous deep learning-based methods fail to comprehensively exploit the spatial semantic and structural information in  deep features in the change detection networks. To resolve these concerns, we first propose a new dataset, HSRW-CD, with a spatial resolution higher than 3 meters for WBCD.  Specifically, it contains a large number of image pairs, widely covering various water body types. Besides, a Spatial Semantics and Continuity Perception (SSCP) attention module is designed to fully leverage both the spatial semantics and structure of deep features in the WBCD networks, significantly improving the discrimination capability for water body. The proposed SSCP has three components: the Multi-Semantic spatial Attention (MSA), the Structural Relation-aware Global Attention (SRGA), and the Channel-wise Self-Attention (CSA). The MSA enhances the spatial semantics of water body features and provides precise spatial semantic priors for the CSA. Then, the SRGA further extracts spatial structure to learn the spatial continuity of the water body. Finally, the CSA utilizes the spatial semantic and structural priors from the MSA and SRGA to compute the similarity across channels. Specifically designed as a plug-and-play module for water body deep features, the proposed SSCP allows integration into existing WBCD models. Numerous experiments conducted on the proposed HSRW-CD and Water-CD datasets validate the effectiveness and generalization of the SSCP. The code of this work and the HSRW-CD dataset will be accessed at https://github.com/QingMa1/SSCP.
\end{abstract}


\end{frontmatter}


\section{Introduction}
Change detection (CD) refers to the technology-driven task, which aims to identify significant changes in land cover or object states by analyzing multi-temporal remote sensing imagery of the same geographic region~\citep{lu2014current}. This task has broad applications in the fields of environment monitoring~\citep{zhou2018change}, resource utilization~\citep{cao2023full}, and disaster evaluation~\citep {gong2015change}. As a critical subfield of CD tasks, Water Body Change Detection (WBCD) is intended to detect the water body surface changes from the image pairs across different periods of the same geographical area, which plays a pivotal role in flood monitoring~\citep{2024flood}, water resource management~\citep{2003water}.

\begin{figure}[htbp]
\centering
\includegraphics[width=0.9\textwidth]{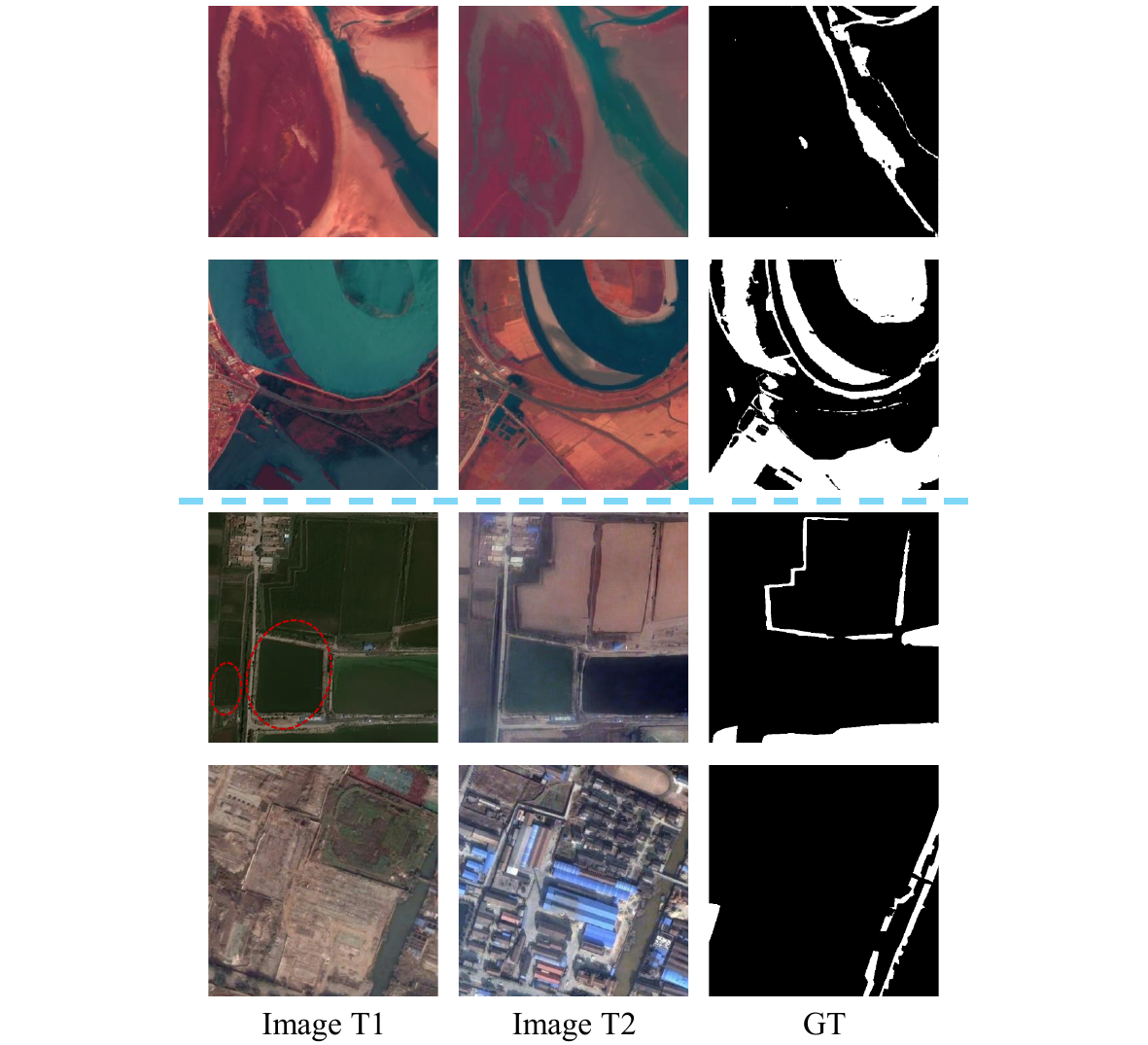}
\caption{The comparison of typical samples in Water-CD and HSRW-CD datasets. The top two rows present examples from the Water-CD dataset, while the following two rows of images come from HSRW-CD. It is clear that the proposed HSRW-CD has a higher spatial resolution than Water-CD. In addition, different water body types vary in shape and have strong spatial continuity.}
\label{water_features}
\end{figure}

With the advancements in deep learning, Convolutional Neural Networks (CNNs) and Vision Transformer (ViT)~\citep{dosovitskiy2020image} have demonstrated excellent performance in WBCD tasks. However, as data-driven tasks, the lack of dedicated datasets for WBCD limits the development of this scientific domain. Therefore, the Water-CD~\citep{li2024interactive} dataset resolves the shortage in publicly accessible datasets for WBCD. However, the images with a spatial resolution of 10 meters in this dataset only cover some relatively large seasonal natural water bodies, such as lakes from the Yangtze River Basin and Jumna River, as illustrated in the top two rows in Fig.~\ref{water_features}. In flood monitoring and water resource management, these images with a spatial resolution of 10 meters may lead to imprecise delineation of flood-affected areas or deviations in reservoir water volume estimation, thereby impacting emergency response and policy-making. Consequently, the Water-CD dataset is inadequate for the management of urban or rural surface water resources. 

Considering inherent characteristics of water bodies presented in images, through suppressing shadow noise in urban areas, the combination of Sentinel-2 Water Index (SWI)~\citep{jiang2021effective} and Otsu algorithm~\citep{otsu1975threshold} achieves better detection performance for water bodies than Normalized Difference Water Index (NDWI)~\citep{rokni2014water}. However, these approaches exhibit limited effectiveness in intricate environmental contexts. MSNANet~\citep{lyu2022multiscale} is proposed to extract water body features at different scales in complex scenarios efficiently, while paying more attention to water body geometric shape characteristics. To tackle the challenge of unclear boundaries and the incorrect classification of shadow regions for water body in high resolution imagery, Sun et al.~\citep{sun2024extraction} propose the WaterDeep, which leverages closely connected ASPP\citep{chen2017rethinking} modules to merge multi-scale information into high-level feature representations, significantly improving the efficiency in differentiating various water bodies in complicated urban background. Nevertheless, deep features of water body images in CD networks contain rich spatial semantic and structural information, which remains underutilized comprehensively in WBCD research. For instance, as illustrated in Fig.~\ref{water_features}, in the third row, the left farmland with high soil moisture in Image T1 is visually similar to the lakes and rivers, which are prone to causing false positives in WBCD. Enhancing spatial semantics in water body deep features can effectively distinguish these different land surface types. In the last row, we can observe that urban waterways have strong spatial continuity. Making use of the spatial structural information contributes to improving the integrity of prediction results in WBCD.

To overcome these issues, we first propose the HSRW-CD dataset, which features a high spatial resolution better than 3 meters. The proposed dataset encompasses 2,085 bi-temporal image pairs, including multiple water body types such as urban waterways, river systems, lakes, and artificial reservoirs. 
As illustrated in Fig.~\ref{water_features}, certain vegetation exhibits visual characteristics similar to water body features. Besides, spatial continuity is also an important characteristic of water bodies. In response to the aforementioned properties of water bodies in remote sensing images, while considering the rich information in deep features, we propose the Spatial Semantics and Continuity Perception (SSCP) attention module to refine deep features in WBCD networks. Specifically, SSCP comprises a Multi-semantic Spatial Attention (MSA), a Structural Relation-aware Global Attention (SRGA), and a Channel-wise Self-Attention (CSA). The MSA first employs spatial semantic attention to derive spatial information across multiple semantic scales, which is from four separate sub-features. 
Then, the MSA-processed feature maps are input into SRGA, which leverages the compact global-scale structural relation patterns to deduce  the attention coefficients. The outputs of MSA and SRGA are integrated through a residual connection~\citep{he2016deep} and subsequently fed into the CSA, which is proposed to alleviate semantic discrepancies across various sub-features in MSA and promote spatial semantic and structural information fusion effectively. Therefore, the proposed SSCP strengthens the spatial semantic and structural information of deep features through the aforementioned steps. 

The primary contributions of the work are enumerated as below:
\begin{enumerate}
\item{Within the scope of our current knowledge, we construct the first high spatial resolution and large-scale dataset HSRW-CD for remote sensing WBCD, featuring imagery with a spatial resolution finer than 3 meters, a collection of 2,085 bi-temporal image pairs, and various water body types. Due to the properties of the HSRW-CD dataset, it will further enhance the application of WBCD in sophisticated water resource management.} 

\item{Focusing on the visual similarity between temporary water bodies and moisture-rich vegetation, as well as the structural continuity of water bodies in remote sensing images, the SSCP attention module is proposed to enhance water body spatial semantic and structural information in deep feature maps. The proposed SSCP is a play-and-plug module, which can be theoretically capable of being integrated into arbitrary CD network architectures.}

\item{Numerous experiments carried out on the newly proposed HSRW-CD and Water-CD datasets validate that the proposed SSCP effectively extracts the features of real water bodies, while improving the spatial continuity of changed regions in WBCD.}
\end{enumerate}

\section{Related work}
\subsection{Water body change detection}

Conventional approaches to detect water body changes include NDWI~\citep{rokni2014water}, SVM~\citep{sarp2017water}, pixel-level image fusion and classification techniques~\citep{rokni2015new}. However, these methods have poor generalization ability, and some methods that require setting appropriate thresholds increase labor and time costs.

In deep learning-based WBCD methods, CNNs and attention mechanisms have been instrumental in advancing the field by virtue of their robust feature extraction and enhancement capabilities, respectively. Lin et al.~\citep{na2023water} propose a WBCD algorithm that combines Siam-U-Net++ and SE attention~\citep{hu2018squeeze} to mitigate the impact of shadow interference in the Three Gorges Reservoir area. To identify tiny water bodies better, MC-WBDN~\citep{yuan2021deep} consolidates three innovative components, effectively leveraging the satellite image data to improve detection performance. In addition, attention mechanisms can also increase the focus on water body features. Cao et al.~\citep{cao2024water} design the novel network EU-Net, which models the spatial contextual relationships by multi-scale attention to improve the discriminative ability for water bodies.

Recently, ViT~\citep{dosovitskiy2020image} has greatly advanced the field of computer vision with its powerful global modeling capabilities. Some researchers have also attempted to apply it in the field of WBCD. Facing the challenges of water body diversities in shape, scale, and reflection characteristics, WaterFormer~\citep{kang2023waterformer} is coupled with the Transformer~\citep{vaswani2017attention} and CNN, thus detecting water bodies in optical high resolution images accurately and efficiently. Inspired by  the time-reversal asymmetry in physics while considering the temporal properties of water bodies, Li et al.~\citep{li2024trsanet} propose the  TRSANet to enhance essential temporal-spatial changed features more effectively for WBCD. Deep features in the neural networks encode rich spatial semantic and structural information. Meanwhile, such information plays a pivotal role in enabling models to understand complex data. However, spatial semantic and structural information in water body deep features of CD networks remains largely unexplored. 

\subsection{Spatial semantics and structural continuity enhancement}

Enhancing the spatial semantic representation of features helps networks to more accurately localize objects, which is mostly achieved through attention mechanisms, specific network architectures, or data augmentation techniques. CBAM~\citep{woo2018cbam} combines the channel and spatial attention mechanisms, in which the spatial weights of the feature maps are dynamically adjusted to refine the spatial semantic features. SKNet~\citep{li2019selective} allows every neuron to adjust the receptive field size adaptively in CNNs in light of different scales of input features, which makes uses of  the SE attention~\citep {hu2018squeeze} to integrate multi-scale spatial information. Li et al.~\citep{li2019spatial} propose a SGE module which aims to reweight each sub-feature to reduce the impact of background noise, while enhancing learned semantic expression and suppressing irrelevant feature representation. In addition,  ViT~\citep{dosovitskiy2020image} encodes the image into sequences by splitting it into patches, and utilizes self-attention to model the global spatial semantics. To explore the combined effects of spatial and channel attention, Si et al.~\citep{si2025scsa} alleviate semantic differences to guide channel recalibration, which effectively enhances discriminative features. Recently, Mamba~\citep{gu2023mamba} proposes selective structured state space models and a novel parallel algorithm, which capture spatial context-dependent information with fast inference. Furthermore, VMamba~\citep{liu2024vmamba} introduces the VSS block,  effectively facilitating global semantic contextual information from various scanning routes.  

The spatial structural continuity information of features plays a crucial role in modeling relative positional relationships and scene topological understanding. Feature pyramid networks~\citep{lin2017feature} enhance spatial structural awareness through multi-scale feature pyramids. Kipf et al.~\citep{kipf2016semi} propose a scalable method to model the topological relationships between nodes through graph convolutional networks. To mine the knowledge from global structure patterns, Zhang et al.~\citep{zhang2020relation} model the pairwise dependencies of all the features for better attention learning. Swin Transformer~\citep{liu2021swin} has the flexibility to capture information at multiple scales, in which the hierarchical architecture models local and global spatial structures via shifted windows.
Moreover, MAE~\citep{he2022masked} proposes a network with asymmetric encoder-decoder for image reconstruction, effectively learning the spatial structural relationships between image patches. Considering the limited representation ability of light-weight CNNs, Chen et al.~\citep{chen2020dynamic} present the dynamic convolution, which fuses multiple parallel convolution kernels according to their attention weights adaptively, enhancing the model's ability to model local spatial structures. MaxViT~\citep{tu2022maxvit} employs blocked global and local range attention to facilitate  spatial interactions between global and local scales across variable input resolutions, simultaneously enhancing multi-scale structural modeling. Inspired by Mamba~\citep{gu2023mamba}, to enhance the flow of contextual visual feature information significantly, Spatial-Mamba~\citep{xiao2024spatial} establishes neighborhood relationship in the state space  directly, leveraging dilated convolutions to learn spatial structural dependencies.

Nevertheless, few studies focus on exploring the synergy between the spatial semantics and spatial structural information in deep features. Meanwhile, the deep features of water body bi-temporal images in CD networks contain rich information. Therefore, this paper aims to effectively combine their spatial semantics and structural information to improve the performance of WBCD.

\section{Method}
\begin{figure*}[hbt!]
\centering
\includegraphics[width=1\textwidth]{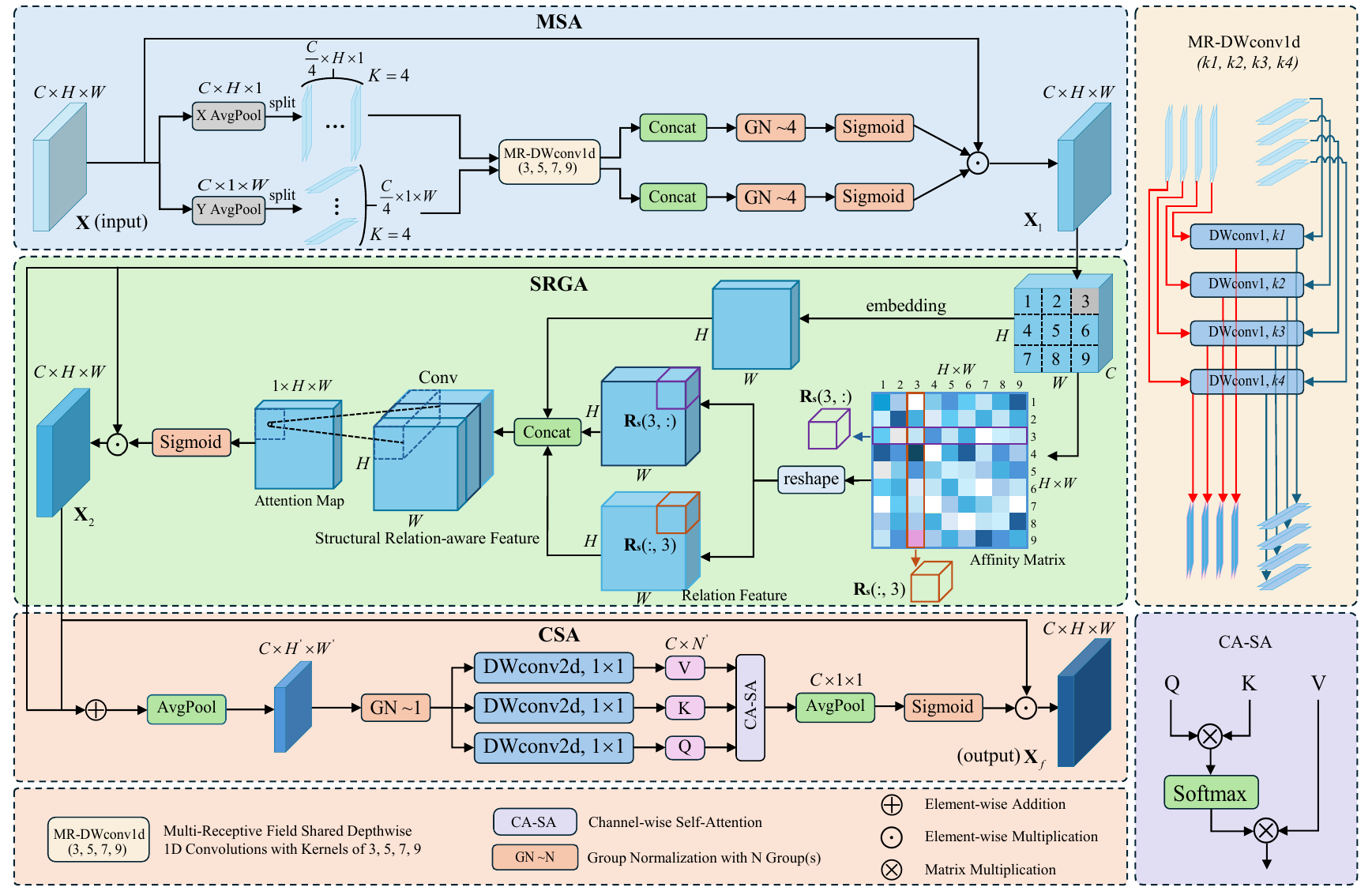}
\caption{The overall structure of the proposed SSCP. Specifically, the deep feature map \textbf{X} from the CD networks passes through the MSA and SRGA module to extract spatial semantics and structural information. Finally, the CSA further deepens understanding by leveraging the spatial semantics and structural priors. }
\label{module}
\end{figure*}

\begin{figure}[hbt!]
\centering
\includegraphics[width=1\textwidth]{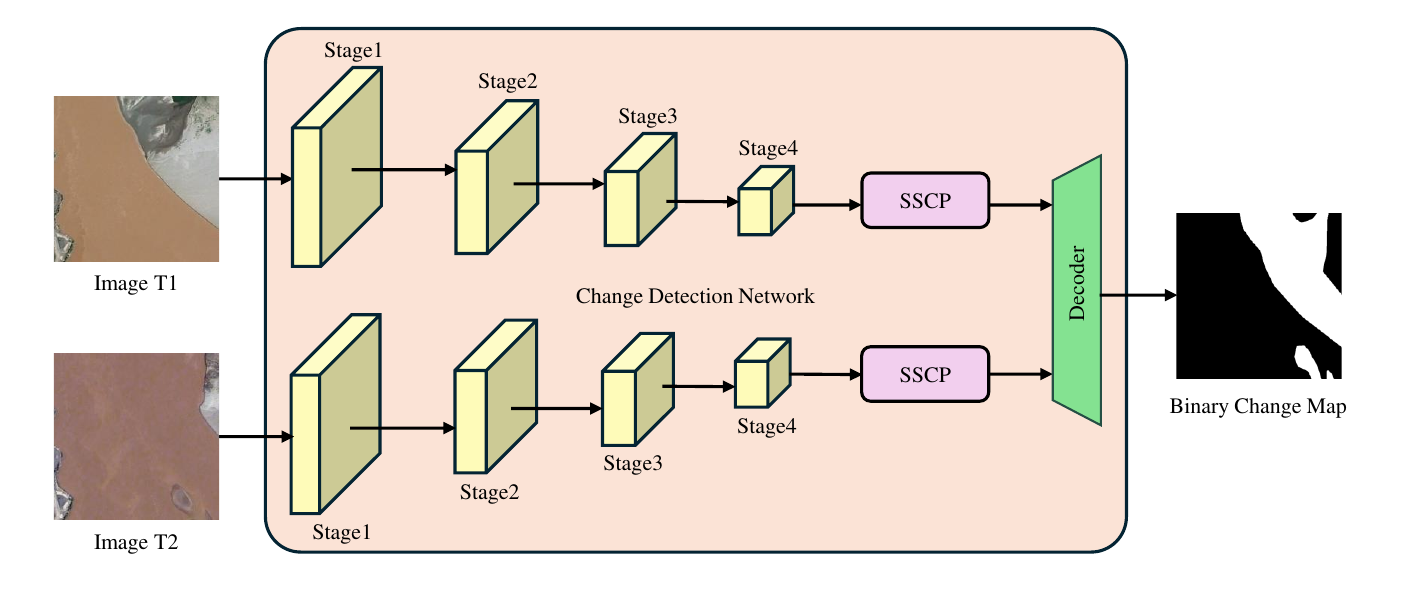}
\caption{The structure of change detection networks with SSCP. The SSCP module is utilized to process deep water body features.}
\label{overview}
\end{figure}

 As illustrated in Fig.~\ref{module}, MSA is employed to enhance spatial semantics in deep features. Subsequently, the SRGA compactly captures the structural relationships. The CSA utilizes the priors from MSA and SRGA to reweight channels. Besides, we can observe that the proposed SSCP attention is  designed to refine deep features in Fig.~\ref{overview}.


\subsection{Multi-semantic spatial attention}

The Multi-semantic Spatial Attention (MSA) aims to further extract spatial semantic information of water bodies from deep features, supply the CSA with accurate spatial semantic priors. 

Specifically, we decompose the input deep feature $\mathbf{X}\in \mathbb{R} ^ {C \times H \times W}$ across the spatial dimensions of height and width using global average pooling. The generated $\mathbf{X}_H\in \mathbb{R} ^ {C \times 1 \times W}$ and $\mathbf{X}_W\in \mathbb{R} ^ {C \times H \times 1}$ are two unidirectional 1D sequences. To grasp different spatial semantic distributions and contextual dependencies, the feature is divided into $K$ equally sized $\mathbf{X}_H^{i}$ and $\mathbf{X}_W^{i}$, where each sub-feature contains $\frac{C}{K}$ channels. The default K value is set to 4. The decomposition into sub-features can be formulated as,
\begin{eqnarray}
\begin{aligned}
        \mathbf{X}_H^{i} &= \mathbf{X}_H[:, (i - 1)\times \frac{C}{K} : i  \times\frac{C}{K}, :, :],\\
        \mathbf{X}_W^{i} &= \mathbf{X}_W[:, (i - 1)\times \frac{C}{K} : i  \times\frac{C}{K}, :, :],
\end{aligned}
\end{eqnarray}
where $\mathbf{X}^{i}$ denotes the $i$-th sub-feature, $i\in[1, K]$. Independent sub-features enable efficient extraction of spatial semantic information.
Subsequently, we intend to extract distinct semantic spatial patterns within every sub-feature. To extract rich semantic information and enhance the semantic coherence of water body deep features, depth-wise 1D convolutions which have 3, 5, 7, and 9 kernel sizes are separately employed on partitioned sub-features. In addition, lightweight shared convolutions are utilized to address the constrained  receptive field arising from decomposition features into $H$ and $W$ dimensions and independent 1D convolution operations. The process for spatial semantic information  extraction is presented as follows,  
\begin{eqnarray}
\begin{aligned}
        \widetilde{\mathbf{X}}_H^{i} &= \text{DWConv1d}_{k_i}^{\frac{C}{K}\rightarrow\frac{C}{K}}(\mathbf{X}_H^{i}), \\
        \widetilde{\mathbf{X}}_W^{i} &= \text{DWConv1d}_{k_i}^{\frac{C}{K}\rightarrow\frac{C}{K}}(\mathbf{X}_W^{i}), \\
\end{aligned}
\end{eqnarray}
where $\widetilde{\mathbf{X}}^{i}$ denotes the spatial semantics in the $i$-th sub-feature generated by light-weighted convolution layers, $\widetilde{\mathbf{X}}_H^{i}\in\mathbb{R} ^ {\frac{C}{K} \times 1\times W }$, $\widetilde{\mathbf{X}}_W^{i}\in\mathbb{R} ^ {\frac{C}{K} \times H \times 1}$. For the sub-feature $i$, $k_i$ signifies the convolutional kernel employed , with $i=1, 2, ..., K$.

Following the decomposition into independent sub-features and the capture of the spatial semantic information at various spatial locations, the spatial semantic attention map is ultimately created. Different semantic sub-features are first concatenated and then  normalized by Group Normalization (GN)~\citep{wu2018group}, which  divides channels into $K$ groups
for normalization. GN enables channel-wise separate  normalization of each sub-feature, avoiding batch variance noise to efficiently alleviate semantic crosstalk among sub-features and prevent attention degradation. Finally, spatial attention weight matrix is derived from the Sigmoid normalization. The output feature can be computed as follows,
\begin{eqnarray}
\begin{aligned}
        \mathbf{A}_H = \sigma (\text{GN}_H^{K}(&\text{Concat}(\widetilde{\mathbf{X}}_H^{1}, \widetilde{\mathbf{X}}_H^{2}, ..., \widetilde{\mathbf{X}}_H^{K}))), \\
        \mathbf{A}_W = \sigma (\text{GN}_W^{K}(&\text{Concat}(\widetilde{\mathbf{X}}_W^{1}, \widetilde{\mathbf{X}}_W^{2}, ..., \widetilde{\mathbf{X}}_W^{K}))),\\
        \text{MSA}(\mathbf{X}) = &\mathbf{X}_1 = \mathbf{X} \cdot \mathbf{A}_H \cdot \mathbf{A}_W,
\end{aligned}
\end{eqnarray}
where $\sigma (\cdot )$ represents the Sigmoid normalization, $\text{GN}_H^{K}(\cdot)$ and $\text{GN}_W^{K}(\cdot)$ stand for GN configured  with $K$ groups across the independent height $(H)$ and width $(W)$ dimensions. The feature $\mathbf{X}_1$ contains enhanced spatial semantic information of the input $\mathbf{X}$.

\subsection{Structural relation-aware global attention}
For modeling long-range pairwise relationships between water body regions, the SRGA module draws on the non-local operation~\citep{wang2018non}, which captures contextual dependencies by computing similarity between all pairs of positions in the feature map. SGRA is proposed to exploit pairwise relations of the present feature node with each feature node. Then these pairwise relations are stacked to denote the overall structure for the current position's node. 

As depicted in Fig.~\ref{module}, for the input feature $\mathbf{X}_1$, the feature vector at each spatial position is 
considered a feature node. Therefore, the graph composed of all positions contains $N=H\times W$ nodes. The spatial positions are scanned and identified as 1, 2, $\cdot\cdot\cdot$, $N$. We define the $N$ feature nodes as $\mathbf{x}_u\in\mathbb{R} ^ {C}$, $u\in[1, N]$. For nodes $u$ and $v$, the pairwise relation \(\mathbf{r}_{u, v}\) is formulated as a dot product to measure affinity in feature representation spaces. The aforementioned process is presented as,
\begin{eqnarray}
\begin{aligned}
        \mathbf{r}_{u, v} = \text{f}_s(\mathbf{x}_u, \mathbf{x}_v) = \varphi_s(\mathbf{x}_u)^T \psi_s(\mathbf{x}_v), 
\end{aligned}        
\end{eqnarray}
where $\varphi_s(\cdot)$ and $\psi_s(\cdot)$ represent two embedding functions, which are realized via the 1 × 1 convolution incorporating  Batch Normalization (BN)~\citep{ioffe2015batch}, with ReLU applied subsequently,
\begin{eqnarray}
\begin{aligned}
    \varphi_s(\mathbf{x}_u) = \text{ReLU}(\mathbf{W}_\varphi\mathbf{x}_u),\\
        \psi_s(\mathbf{x}_u) = \text{ReLU}(\mathbf{W}_\psi\mathbf{x}_u),
\end{aligned}        
\end{eqnarray}
where $\mathbf{W}_\varphi\in \mathbb{R}^{\frac{C}{p_1}\times C}$,  $\mathbf{W}_\psi\in \mathbb{R}^{\frac{C}{p_1}\times C}$. A predefined positive integer ${p_1}$ is used to regulate the dimension reduction ratio. Similarly, the affinity from node $v$ to node $u$ is obtained as $\mathbf{r}_{v,u}=\text{f}_s(\mathbf{x}_v, \mathbf{x}_u)$. The pair $(\mathbf{r}_{u, v}, \mathbf{r}_{v, u})$ indicates bidirectional relations between feature nodes $\mathbf{x}_u$ and $\mathbf{x}_v$. An affinity matrix $\mathbf{R}_s\in \mathbb{R}^{N\times N}$ denotes all the pairwise relations among nodes. Then, its pairwise relations with every node are stacked in the raster scan order for the $u$-th feature node. Therefore, a relation vector $\mathbf{r}_u = [\mathbf{R}_s(u, :), \mathbf{R}_s(:, u)] \in \mathbb{R}^{2N}$ is obtained. As shown in Fig~\ref{module}, we take row $3$ and column $3$ of $\mathbf{R}_s$ as an example, its relation vector $\mathbf{r}_3 = [\mathbf{R}_s(3, :), \mathbf{R}_s(:, 3)]$ is utilized as the relational feature to compute the attention of the third spatial position.

Besides, both the global-scale  structure relevant to the feature $\mathbf{x}_u$ itself and the local primitive   information are considered. As the above two pieces of information belong to different feature spaces, we separately embed and concatenate them for obtaining the structural continuity-aware feature $\widetilde{\mathbf{y}}_u$,
\begin{eqnarray}
\begin{aligned}
\mathbf{\theta}_s(\mathbf{x}_u) = &\text{ReLU}(\mathbf{W}_\theta\mathbf{x}_u),\\
\eta_s(\mathbf{r}_u) = &\text{ReLU}(\mathbf{W}_\eta\mathbf{r}_u),\\
\widetilde{\mathbf{y}}_u=\text{Concat}(&\text{Pool}_c(\theta_s(\mathbf{x}_u),\eta_s(\mathbf{\mathbf{r}}_u))),
\end{aligned}
\end{eqnarray}
 where $\theta_s$ and $\eta_s$ represent the embedding transformations for the feature representation  and its contextual relational information. Similarly, these transformations are also achieved via the $1\times1$ convolution and BN with ReLU function. $\mathbf{W}_\theta\in\mathbb{R}^{\frac{C}{p_1}\times C}$, $\mathbf{W}_\eta\in\mathbb{R}^{\frac{2N}{2p_1}\times2N}$. $\text{Pool}_c(\cdot)$ signifies the global channel-wise average pooling, which further compresses the dimension to 1, and $\widetilde{\mathbf{y}}_u\in\mathbb{R}^{1+N/p_1}$.

Note that the global relations contain rich spatial structure features, the valuable knowledge from them is mined to infer attention weights by a trainable model. For the $u$-th feature, its spatial attention $\mathbf{a}_u$  is acquired using a modeling function as follows,
\begin{eqnarray}
\begin{aligned}
        \mathbf{a}_u=\sigma(\mathbf{W}_2\text{ReLU}(\mathbf{W}_1\widetilde{\mathbf{y}}_u)), \\
        \mathbf{a}=\text{Concat}(\mathbf{a}_1, \mathbf{a}_2, ...,\mathbf{a}_N),\\
        \text{SRGA}(\mathbf{X}_1)=\mathbf{X}_2=\mathbf{X}_1\cdot{\mathbf{a}},
\end{aligned}
\end{eqnarray}
where $\mathbf{W}_1$ and $\mathbf{W}_2$ are both composed of a $1\times1$ convolution with BN applied afterward. $\mathbf{W}_1$ reduces the channel dimension by a factor of $p_2$, and $\mathbf{W}_2$ further projects it to 1. $\sigma(\cdot)$ represents the Sigmoid normalization, $\mathbf{a}$ denotes the final attention map. The output $\mathbf{X}_2$ represents spatial continuity-enhanced $\mathbf{X}_1$.

\subsection{Channel-wise self-attention}
To fully utilize the spatial semantic and structural priors from MSA and SRGA, respectively, we propose CSA to further measure the similarity across the channels in the input feature $\mathbf{X}_3$. Specifically, $\mathbf{X}_3$ is obtained by element-wise addition of features $\mathbf{X}_1$ and $\mathbf{X}_2$, which preserves multi-scale semantics while reinforcing structural consistency.

Motivated  by the notable merits of ViT~\citep{dosovitskiy2020image} in leveraging multi-head self-attention for modeling spatial dependencies among distinct  tokens, the single-head self-attention and spatial prior information from MSA and SRGA are combined to measure inter-channel similarities. Besides, to further maintain  and exploit the spatial semantic and structural information from MSA and SRGA, we adopt a stepwise  compression approach using average pooling, effectively reducing the loss of essential  feature information.Motivated  by the notable merits of ViT~\citep{dosovitskiy2020image} in leveraging multi-head self-attention for modeling spatial dependencies among distinct  tokens, the single-head self-attention and spatial prior information from MSA and SRGA are combined to measure inter-channel similarities. Channel-wise recalibration in  CSA  is inspired by the Squeeze-and-Excitation module~\citep{hu2018squeeze}. To further maintain  and exploit the spatial semantic and structural information from MSA and SRGA, we adopt a stepwise  compression approach using average pooling, effectively reducing the loss of essential  feature information. The detailed process of CSA is as follows,
\begin{eqnarray}
    \begin{aligned}
        \mathbf{X}_3 &= \mathbf{X}_1 + \mathbf{X}_2,\\
        \mathbf{X}_{p} &= \text{Pool}_{(7, 7)}^{(H, W)\rightarrow(H^{\prime}, W^{\prime})}(\mathbf{X}_3),\\
        \mathbf{F}_{proj} &= \text{DWConv1d}_{(1, 1)}^{C\rightarrow C},\\
        \mathbf{Q} &= {\text{F}_{proj}^{Q}(\mathbf{X}_p)},\\
        \mathbf{K} &= {\text{F}_{proj}^{K}(\mathbf{X}_p)},\\
        \mathbf{V} &= {\text{F}_{proj}^{V}(\mathbf{X}_p)},\\
        \mathbf{X}_{attn}=\text{Attn}(&\mathbf{Q},\mathbf{K},  \mathbf{V})=\text{Softmax}(\frac{\mathbf{Q} \mathbf{K}^{\text{T}}}{\sqrt{C}})\mathbf{V}, \\
        \text{CSA}(\mathbf{X}_3)= \mathbf{X}_f=&\mathbf{X}_3 \cdot \sigma(\text{Pool}_{(H^{\prime}, W^{\prime})}^{(H^{\prime}, W^{\prime})\rightarrow(1, 1)}(\mathbf{X}_{attn})),
    \end{aligned}
\end{eqnarray}
where the pooling operation denoted by $\text{Pool}_{(k, k)}^{(H, W)\rightarrow(H^{\prime}, W^{\prime})}(\cdot)$ employs a $k \times k$ kernel, which performs resolution rescaling from $(H, W)$ to $(H^\prime, W^\prime)$. $\text{F}_{proj}(\cdot)$ stands for the linear transformation in self-attention. $\mathbf{X}_f$ is the enhanced feature derived from the original feature $\mathbf{X}$ through spatial semantic and structural continuity enhancement, as well as channel-wise reweighting.
\section{Results and analysis}
The Water-CD and the proposed HSRW-CD datasets are first introduced. Subsequently, the setup of our experiments is presented. Then, the effectiveness and efficiency of the proposed SSCP for WBCD are verified through extensive experiments. Furthermore, ablation and analysis studies are conducted to validate the functionality of the SSCP's components .

\subsection{Datasets}
To fully prove the effectiveness of the  SSCP attention module proposed in WBCD tasks, our experiments are conducted on two distinct datasets, which are described as follows.

\textbf{Water-CD~\citep{li2024interactive}} fills a gap in publicly available datasets for WBCD. The dataset includes seasonal lakes from the Yangtze River Basin in China and various seasonal water bodies from the Jumna River in South Asia. These image pairs comprise cloud-free data captured during both rainy and dry seasons. Utilizing Sentinel-2 satellite imagery, it includes 1,149 bi-temporal image pairs with a 10-meter spatial resolution, each sized at 512 $\times$ 512. We adhere to the official data split, which has 765 pairs for training, 207 pairs for validation, and 177 for testing.

\begin{figure*}[hbt!]
\centering
\includegraphics[width=1\textwidth]{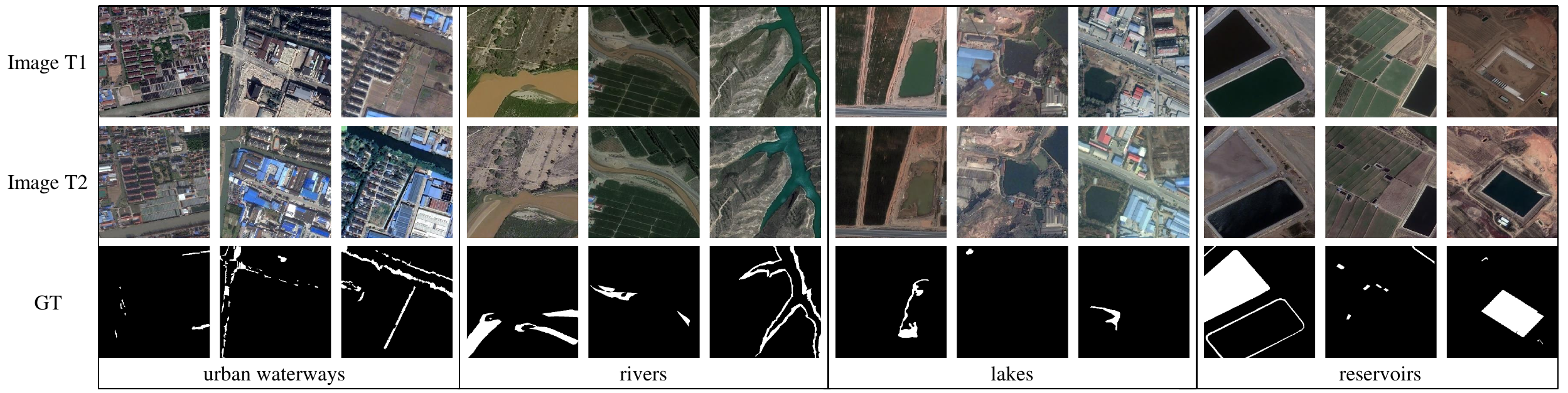}
\caption{Samples of various water body types from the proposed HSRW-CD dataset. Each group of three consecutive columns represents sample images of a specific water body type from left to right.}
\label{samples}
\end{figure*}

\textbf{The proposed HSRW-CD dataset} comprises 2,085 image pairs spanning diverse geographical regions and environmental contexts, encompassing various water body types such as urban waterways, river systems, lacustrine environments, and artificial reservoirs. These image pairs are from more than 7 cities or districts in China, including Beijing, Chongqing, Chengdu, Hangzhou, Wuhan, Shenzhen, and Shanghai. The bi-temporal images are manually collected from high-resolution satellites and Google Earth. The annotation of the dataset is carried out by an expert group of Earth vision applications, which guarantees high label accuracy. In addition, the bi-temporal images have a high spatial resolution finer than 3 meters, each sized at 512 $\times$ 512 pixels. Following a 7:1:2 proportion, the image dataset is randomly divided  into training, validation, and test subsets according to a 7:1:2 proportion, leading to training, validation, and test sets containing 1,476, 203, and 406 independent image pairs. Examples from the HSRW-CD dataset are illustrated in Fig.~\ref{samples}. Notably, the proposed dataset encompasses diverse water body types.

\subsection{Experimental configuration}

Specifically, the experiments are performed on the RTX 4090. During the training period, the network is optimized using the AdamW optimizer. The learning rate is 1e-4, and so is the weight decay. The training configuration specifies a batch size of 8. We train on the HSRW-CD and Water-CD for 40,000 iterations.

\subsection{Evaluation metrics}
For the performance assessment of WBCD approaches, following evaluation metrics are applied to our experiments. Specifically, the ratio of correctly classified samples is reflected by Overall Accuracy (OA). The Precision (Pre.) represents the correctness  of all predicted positives. The Recall (Rec.) is used to evaluate the ratio of positive samples in all correct predictions. The Intersection over Union (IoU) quantifies the spatial overlap ratio between the inferred changed pixels and the ground-truth changed pixels from a geometric overlap perspective. The F1-score (F1) functions as an balanced indicator of  recall and precision. The mathematical formulations of these metrics are presented below,
\begin{eqnarray}
    \text{OA} = \frac{\text{TP + TN}}{\text{TP + FN + TN + FP}},
\end{eqnarray}

\begin{eqnarray}
    \text{Pre}. = \frac{\text{TP}}{\text{TP + FP}},
\end{eqnarray}

\begin{eqnarray}
    \text{Rec}. = \frac{\text{TP}}{\text{TP + FN}},
\end{eqnarray}

\begin{eqnarray}
    \text{F1} = \frac{2 \times \text{Pre.} \times \text{Rec.}}{\text{Pre}. + \text{Rec}.},
\end{eqnarray}

\begin{eqnarray}
    \text{IoU} = \frac{\text{TP}}{\text{TP + FN + FP}}.
\end{eqnarray}
\textbf{Note that the F1 and IoU serve as the primary metrics for evaluating WBCD performance}.

\subsection{Results}

\begin{table*}
    \centering
    \tiny
    \caption{The test set-based performance comparison of WBCD tasks on the proposed HSRW-CD and Water-CD datasets between the state-of-the-art approaches and certain approaches after integrating SSCP. Where SSCP\underline{~}BIT denotes the BIT method with SSCP. \textcolor{red}{Red} and \textcolor{blue}{blue} are used to highlight the optimal and sub-optimal results, respectively.}
    \label{all_methods_table1}
    \renewcommand{\arraystretch}{1.5}
    \setlength{\tabcolsep}{3pt}
    \begin{tabular}{lllccccccccccccc}
        \toprule
        \multirow{2}{*}{Method} &\multirow{2}{*}{Year} & \multirow{2}{*}{Backbone} & \multicolumn{5}{c}{HSRW-CD} & \multicolumn{5}{c}{Water-CD} \\ \cline{4-13}
        & & & F1(\%) & Pre.(\%) & Rec.(\%) & IoU(\%) & OA(\%)
        & F1(\%) & Pre.(\%) & Rec.(\%) & IoU(\%) & OA(\%) \\
        \midrule
        BIT~\citep{chen2021bit}  & 2021& ResNet18 & 70.03 & 79.82 & 62.38 & 53.88 & 98.60 &  90.66 & 92.33 & 89.06 & 82.92 & 96.64 \\
        SNUNet~\citep{fang2021snunet}& 2022 & NestedUNet & 62.07 & 70.72 & 55.31 & 45.00 & 98.22 & 89.67 & 89.02 & 90.33 & 81.27 & 96.18\\
        ChangeFormer\citep{changeformer} & 2022& MiT-B0 & 74.13 & 83.41 & 66.70 & 58.89 & 98.78 & 90.78
        & 91.81 & 89.78 & 83.12 & 96.66\\
        LightCDNet-L\citep{yang2023lightcdnet} & 2023& ShuffleNetV2 & 60.48 & 68.48 & 54.15 & 43.35 & 98.14 & 89.73 & 90.27 & 89.20 & 81.38 & 96.26  \\
        HANet\citep{han2023hanet} & 2023& HAN & 59.92 & 68.82 & 53.06 & 42.77 & 98.13 & 89.92 & 90.34 & 89.49 & 81.68 & 96.32\\
        Changer\citep{fang2023changer} & 2023& ResNeST-101 & 68.70 & 74.38 & 63.83 & 52.33 & 98.47 & 90.58 & 92.10 & 89.11 & 82.78 & 96.60 \\
        CGNet\citep{han2023change} & 2023& VGG-16 & 71.03 & 78.56 & 64.81 & 55.07 & 98.61 & 90.75 & 91.73 & 89.79 & 83.06 & 96.64  \\
        DDLNet~\citep{ma2024ddlnet} & 2024& ResNet18 & 69.72 & 78.18 & 62.92 & 53.52 & 98.56 & 89.90 & 91.31 & 88.53 & 81.65 & 96.35\\
        CDMaskFormer~\citep{ma2024cdmaskformer} & 2024& SeaFormer-B & 74.17 & 77.14 & \textcolor{red}{71.42} & 58.95 & 98.69  & 90.71 & 92.25 & 89.22 & 83.00 & 96.65\\
        BAN~\citep{li2024ban} & 2024& MiT-B0 & 74.63 & 83.90  & 67.20  & 59.53 & 98.80 & 91.41  & 92.25 & 90.59 & 84.18 & 96.88\\
        BAN~\citep{li2024ban} & 2024& MiT-B2 & 76.47 & \textcolor{blue}{86.30}  & 68.66  & 61.91 & \textcolor{blue}{98.89} & \textcolor{blue}{91.80} & \textcolor{blue}{92.83} & 90.80 & \textcolor{blue}{84.85} & \textcolor{blue}{97.03}  \\
        CDXLSTM~\citep{CDXLSTM} & 2025& SeaFormer-B & 70.73 & 76.93 & 65.46 & 54.72 & 98.58 & 89.85 & 91.33 & 88.42 & 81.58 & 96.34\\
        \hline        
        SSCP\underline{~}BIT& - & ResNet18 & 72.49 & 81.96 & 64.97 & 56.85 & 98.70 & 90.81 & 92.18 & 89.49 &  83.17 & 96.68 \\
        SSCP\underline{~}ChangeFormer & -& MiT-B0 & 75.84 & 84.05 & 69.10 & 61.09 & 98.84  & 90.94 & 91.97 & 89.93 &  83.38 & 96.71\\
        SSCP\underline{~}CGNet& - & VGG-16  & 71.92 & 81.54 & 64.32 & 56.15 & 98.68 & 91.25  & 92.75 & 89.81 & 83.92& 96.84\\
        SSCP\underline{~}DDLNet& -  & ResNet18  & 70.88 & 71.92 & \textcolor{blue}{69.87} & 54.89 & 98.49 & 90.13  & 91.85 & 88.47 & 82.03 & 96.45\\
        SSCP\underline{~}CDXLSTM& -  & SeaFormer-B & 72.42 & 76.79 & 68.53 & 56.77 & 98.63 & 89.97 & 91.63 & 88.37 & 81.77 & 96.39 \\
        SSCP\underline{~}BAN & - & MiT-B0 & \textcolor{blue}{76.50} & 84.61 & 69.80 & \textcolor{blue}{61.94} & 98.87 & 91.61 & 92.27 & \textcolor{blue}{90.95} &  84.52 & 96.94  \\
        SSCP\underline{~}BAN & - & MiT-B2 & \textcolor{red}{77.11} & \textcolor{red}{87.38} & 69.00 & \textcolor{red}{62.75} & \textcolor{red}{98.92} & \textcolor{red}{92.07} & \textcolor{red}{92.92} & \textcolor{red}{91.22} &  \textcolor{red}{85.30} & \textcolor{red}{97.12}
        \\
        \bottomrule
\end{tabular}   
\end{table*}

This section presents  comprehensive experimental results on two datasets dedicated to WBCD. The comparison between the proposed method and other cutting-edge approaches is shown in the following content. Besides, the generalization of the proposed SSCP to the mainstream models is validated.   

\textbf{HSRW-CD}. The quantitative metrics of the proposed approach and other advanced models are illustrated in Table~\ref{all_methods_table1}. It can be observed that SSCP\underline{~}BAN with MiT-B2 backbone achieves 77.11\% F1 and 62.75\% IoU, which outperforms other models. Specifically, the SSCP\underline{~}BAN with MiT-B2 backbone surpasses the corresponding BAN method by 0.64\% F1 and 0.84\% IoU, respectively. Besides, to verify the efficacy of the proposed attention, the experiments that integrate the SSCP into some mainstream approaches are designed. The comparison results of the experiments reveal that these methods leveraging the SSCP significantly improve the performance of the original those. For instance, the SSCP\underline{~}BIT is 2.46\% and 2.97\% higher in F1 and IoU than the BIT separately. The SSCP\underline{~}CDXLSTM improves 1.69\% F1 and 2.05\% IoU over the CDXLSTM. Considering that HSRW-CD covers diverse types of water bodies in environmental contexts and has large-scale image pairs, we believe the SSCP can effectively adapt to the features of water bodies to boost WBCD. The results on the Water-CD also support the assumption.

\begin{figure*}
\centering
\includegraphics[width=1\textwidth]{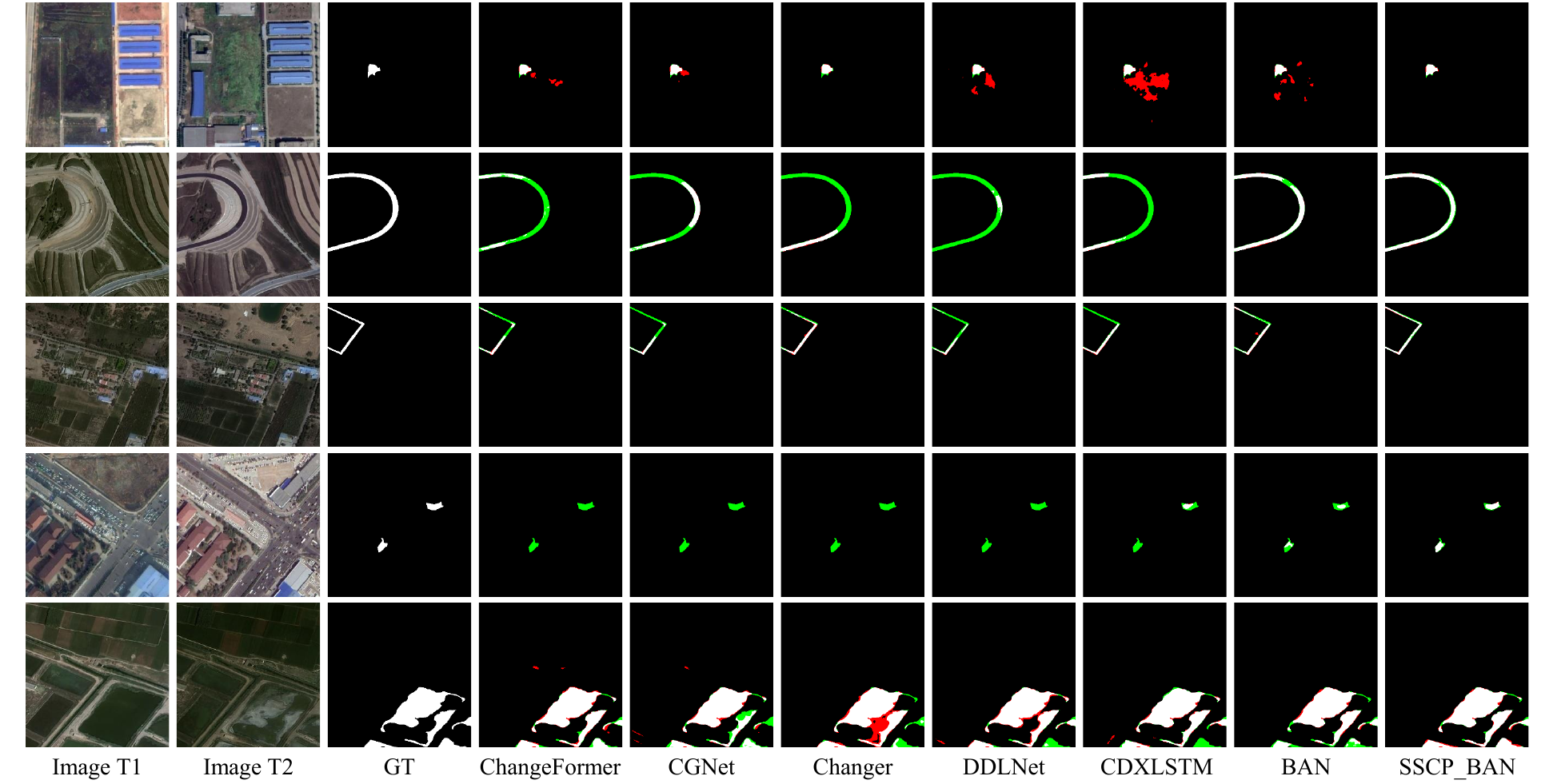}
\caption{The test set visualization results of the HSRW-CD dataset compared with other advanced approaches. The red regions denote the false positive and the green regions are the false negative.}
\label{HSRW-CD_results}
\end{figure*}

\begin{figure*}
\centering
\includegraphics[width=1\textwidth]{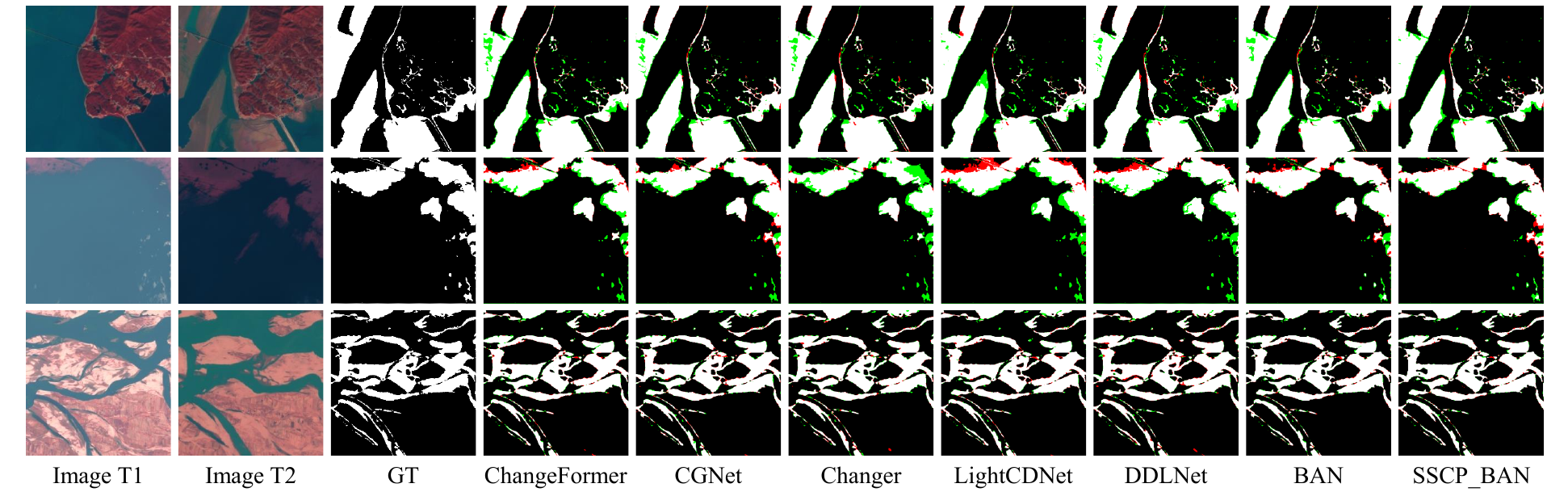}
\caption{The test set visualization results of the Water-CD dataset compared with other state-of-the-art models. The red regions are the false positive and the green regions denote the false negative.}
\label{Water-CD_results}
\end{figure*}
In order to more intuitively display the advantages of the proposed approach, some test set results of SSCP\underline{~}BAN with MiT-B2 and other leading methods are visualized. Corresponding visual results are presented in Fig.~\ref{HSRW-CD_results}. Through the enhancement of spatial semantic and structural information in deep features of water body images, we can observe that the proposed attention effectively adapts to the properties of water bodies. Compared to other methods, the prediction results for changed regions of the SSCP\underline{~}BAN are more complete. Take the top row as an example, the prediction of SSCP\underline{~}BAN has fewer omissions and false detections for detecting the disappeared irregular lake, though it bears visual resemblance to environmental contexts.      

\textbf{Water-CD~\citep{li2024interactive}}. Extensive experiments are also carried out to compare state-of-the-art models on  Water-CD dataset. The data  in Table~\ref{all_methods_table1} show that the SSCP\underline{~}BAN with MiT-B2 achieves 92.07\% F1 and 85.30\% IoU, which performs better than the other methods. It also elevates 0.27\% F1 and 0.45\% IoU, higher than the second BAN, respectively. The SSCP\underline{~}BIT surpasses 0.15\% F1 and 0.25\% IoU over the origin BIT. The effectiveness of the SSCP is verified by these Water-CD experiments.

As shown in Fig.\ref{Water-CD_results}, the visual results on test set of HSRW-CD are presented. As observed,  the predictions of SSCP\underline{~}BAN with MiT-B2 outperform other advanced methods. For instance, in the second row, SSCP\underline{~}BAN predicts fewer false positive regions and omissions, though Image T2 exhibits relatively dark imaging. Though enhanced semantics and continuity extraction of the proposed SSCP for deep features, the model fusing SSCP improve the capacity for identifying water body. As Water-CD dataset is composed of seasonal water bodies from various districts, these quantitative and visual results indicate the generalization of the SSCP on this type of water body.

To further demonstrate the robustness the proposed SSCP, independent-samples T-test are conducted, and we present confidence intervals or standard deviations. The results are shown in Table~\ref{significance_test}. It is evident that the effect of the proposed SSCP is robust. Even if the improvement in model performance is not substantial on Water-CD dataset, the impact of the SSCP attention is still significant($\alpha$$<$0.05). Besides, the standard deviations of F1 scores based on SSCP\underline{~}BIT are lower than those based on BIT~\citep{chen2021bit} baseline across  two datasets, indicating the performance improvement is more stable.

\begin{table}
    \centering
    \scriptsize
     \caption{The results of independent-samples T-tests, standard deviations and confidence intervals targeting F1. Where SD1 denotes the standard deviations of baseline, SD2 represents the standard deviations of the SSCP-integrated baseline, DF signifies degrees of freedom. }
    \label{significance_test}
    \renewcommand{\arraystretch}{1.5}
    \setlength{\tabcolsep}{2pt}
    \scriptsize
    \begin{tabular}{cllccccccc}
        \toprule
        Dataset & Method & Backbone & SD1 & SD2 & 95\% CI(\%)  & T-value & P-value  & DF & Significance ($\alpha$=0.05) \\
        \midrule
         \multirow{2}{*}{HSRW-CD}& BAN~\citep{li2024ban} & MiT-B0 &0.0051& 0.0060& [0.21, 0.85]
        & 3.3654 & 0.0015 & 48.00 & significant \\
        &BIT~\citep{chen2021bit}&ResNet18& 0.0053 & 0.0035& [0.83, 1.87] & 5.6436 & 0.0001 & 12.00 & significant \\
        \multirow{2}{*}{Water-CD}  & BAN~\citep{li2024ban}& MiT-B0 &0.0005& 0.0010& [0.02, 0.20]
        & 2.7261 & 0.0215 & 9.89 & significant \\
        &BIT~\citep{chen2021bit}&ResNet18& 0.0015 & 0.0008&[0.09, 0.37] & 3.6312 & 0.0001 & 12.00 & significant \\
        \bottomrule
    \end{tabular}
\end{table}
\subsection{Ablation and analysis studies}
\textbf{Overall SSCP}. For the demonstration of the effectiveness of the SSCP proposed further, ablation studies of overall SSCP are conducted on HSRW-CD.  Table~\ref{some_methods_w/o_SSCP_table} shows quantitative ablation results. We can observe that SSCP\underline{~}BAN with MiT-B0 is 1.87\% F1 and 2.41\% IoU higher than original BAN, respectively, while only introducing an increase of 0.04M model parameters and 0.02G floating-point computational operations. In addition, SSCP\underline{~}CDXLSTM achieves 1.69\% F1 and 2.05\% IoU improvement over CDXLSTM with an increase of 0.02M parameters and 0.01G computational operations. Notably, most of the methods with SSCP introduce a small increase in model parameters and computational complexity, significantly improving the performance. Meanwhile, after incorporating the SSCP attention, these models maintain inference speeds comparable to those without the attention, which also demonstrates the efficiency of the proposed SSCP.
\begin{figure*}[hbt!]
\centering
\includegraphics[width=1\textwidth]{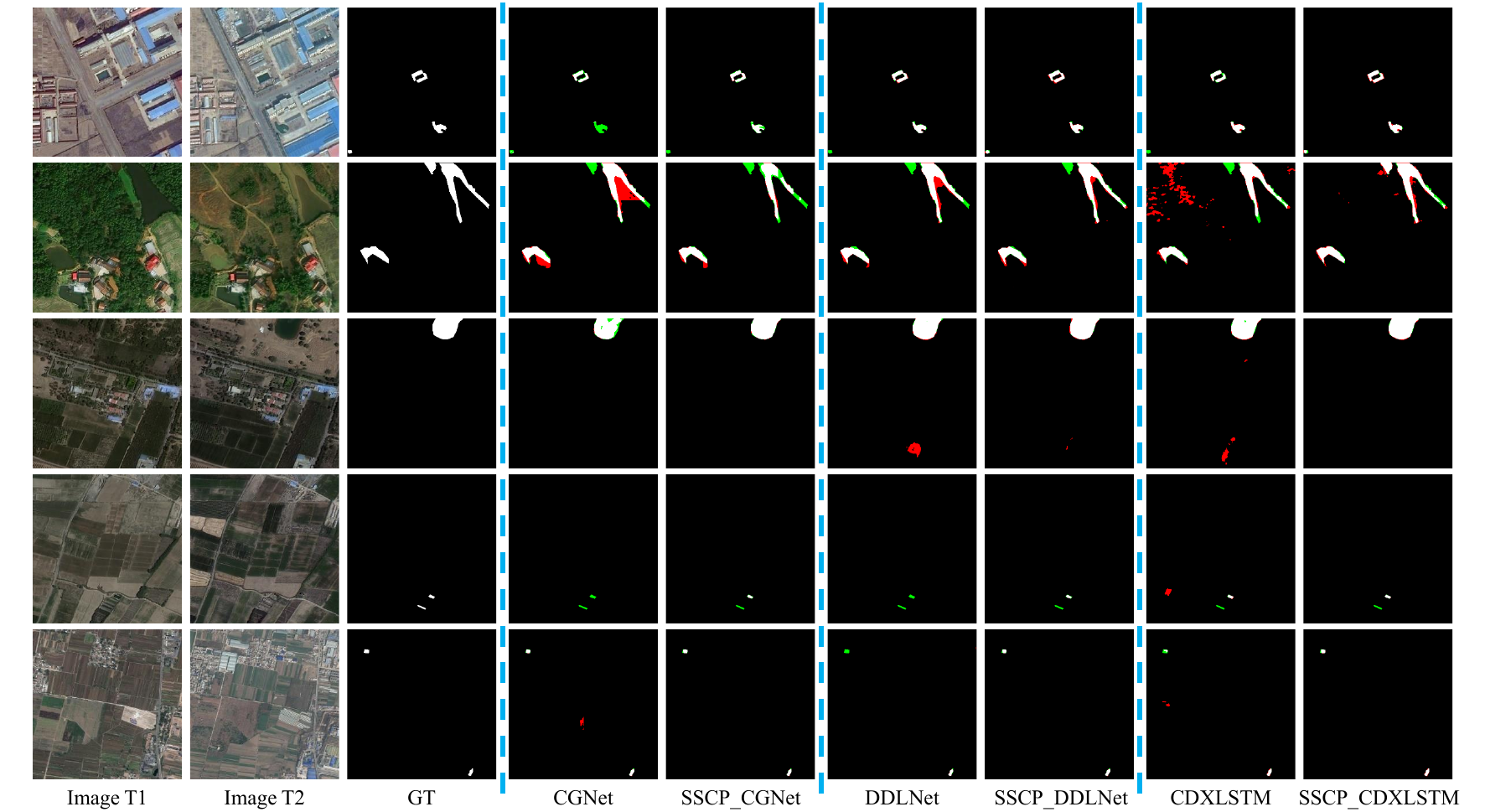}
\caption{The HSRW-CD test set visualization results for overall SSCP ablation. The red regions represent the false positive, while the green regions are the false negative.}
\label{module_ablation}
\end{figure*}
\begin{figure*}[hbt!]
\centering
\includegraphics[width=1\textwidth]{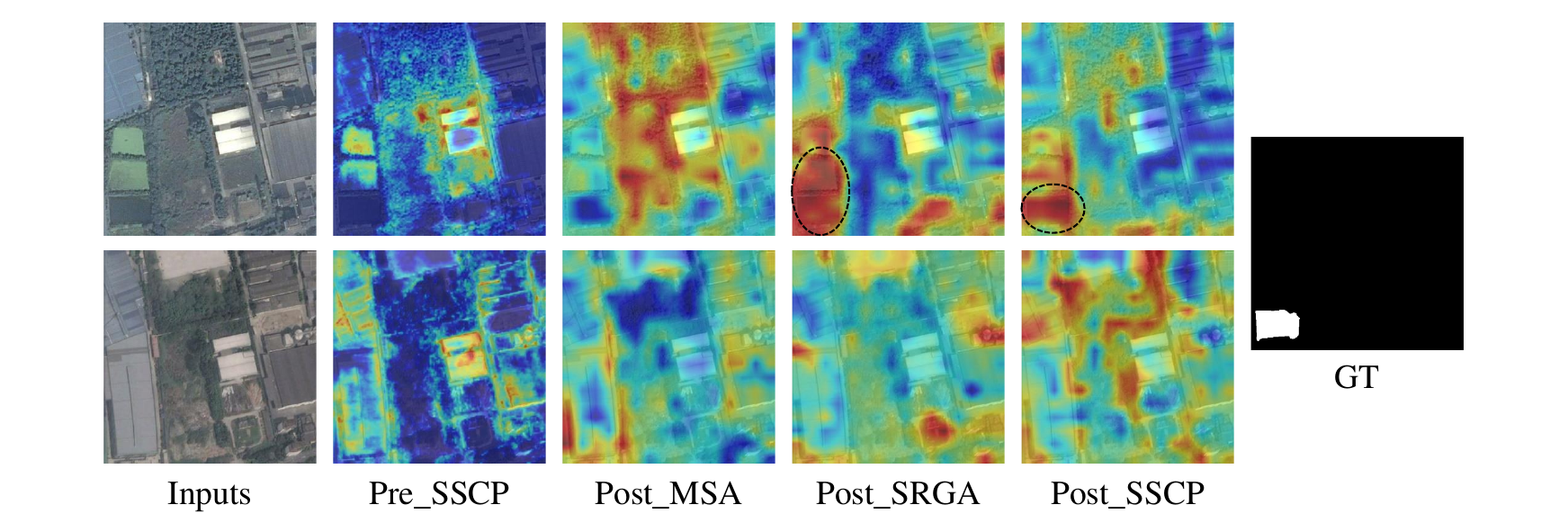}
\caption{The visualization of feature maps from SSCP\underline{~}BAN with backbone MiT-B0. The input bi-temporal image pair is from the HSRW-CD test set. Pre\underline{~}SSCP and Post\underline{~}SSCP denote the feature maps before and after passing through the SSCP, respectively. Post\underline{~}MSA and Post\underline{~}SRGA represent the feature maps processed by the MSA and SRGA sequentially. The red regions represent high activation values, while blue regions represent low activation values.}
\label{BAN-B0_feature_maps_vis}
\end{figure*}
The visual results are also presented in Fig.~\ref{module_ablation}. For instance, the SSCP\underline{~}CGNet denotes the CGNet integrating the SSCP module. It is obvious that the methods with SSCP capture more precise changes. The prediction results in the third row indicate that the SSCP improves the coherence of changed regions, while suppressing the expression of misleading features significantly. Additionally, in order to more intuitively illustrate the efficacy of the proposed SSCP, the feature maps are visually shown in Fig.~\ref{BAN-B0_feature_maps_vis}. It can be observed that the SSCP reallocates the weight of features in images in the network. The final changed regions are allocated with higher activation values in the heatmap SSCP-processed. These visual experimental results further validate the positive impacts of the SSCP for water body detection.

\begin{table*}[htb]
    \centering
    \scriptsize
    \caption{The ablation comparison of model parameters and efficiency on HSRW-CD dataset between some state-of-the-art methods and these methods after integrating SSCP.}
    \label{some_methods_w/o_SSCP_table}
    \renewcommand{\arraystretch}{1.2}
    \setlength{\tabcolsep}{3pt}
    \footnotesize
    \begin{tabular}{llllcll}
        \toprule
        \multirow{2}{*}{Method} & \multirow{2}{*}{Backbone} &   \multirow{2}{*}{Params(M)} &  \multirow{2}{*}{FLOPs (G)} & \multirow{2}{*}{FPS(img/s)}&\multicolumn{2}{c}{HSRW-CD}  \\ \cline{6-7}
        & &&&& F1(\%) & IoU(\%)  \\ 
        \midrule
        
        ChangeFormer\citep{changeformer} & MiT-B0  & 3.85 & 11.38 & 39.71 & 74.13 & 58.89  \\
        SSCP\underline{~}ChangeFormer  & MiT-B0  & 3.88$_{(+0.03)}$ & 11.40$_{(+0.02)}$ & 39.45 &75.84$_{(\uparrow 1.71)}$ & 61.09$_{(\uparrow 2.20)}$ \\
        \hline
        CGNet\citep{han2023change} & VGG-16  & 38.99 & 87.55 & 10.81& 71.03 & 55.07  \\
        SSCP\underline{~}CGNet  & VGG-16  & 39.11$_{(+0.12)}$ & 87.62$_{(+0.07)}$ & 9.83 &71.92$_{(\uparrow 0.89)}$ & 56.15$_{(\uparrow 1.08)}$ \\
        \hline
        DDLNet~\citep{ma2024ddlnet} & ResNet18 & 13.78 & 29.40  & 10.96 & 69.72 & 53.52 \\
        SSCP\underline{~}DDLNet & ResNet18  & 13.84$_{(+0.06)}$ & 29.44$_{(+0.04)}$ & 10.74& 70.88$_{(\uparrow 1.16)}$ & 54.89$_{(\uparrow 1.37)}$ \\
        \hline        CDXLSTM~\citep{CDXLSTM} & SeaFormer-B & 16.19 & 16.88 & 8.14  & 70.73 &  54.72\\
        SSCP\underline{~}CDXLSTM &  SeaFormer-B  & 16.21$_{(+0.02)}$ & 16.89$_{(+0.01)}$ & 8.07 & 72.42$_{(\uparrow 1.69)}$ & 56.77$_{(\uparrow 2.05)}$  \\
        \hline
        BAN~\citep{li2024ban}  & MiT-B0  & 232.07 & 133.97 & 17.88& 74.63 & 59.53 \\
        SSCP\underline{~}BAN & MiT-B0  & 232.11$_{(+0.04)}$ & 133.99$_{(+0.02)}$ & 17.77 & 76.50$_{(\uparrow 1.87)}$ & 61.94$_{(\uparrow 2.41)}$ \\
        \hline
        BAN~\citep{li2024ban} & MiT-B2  & 254.66 & 170.21 & 13.81 & 76.47 & 61.91   \\
        SSCP\underline{~}BAN & MiT-B2  & 254.73$_{(+0.07)}$& 170.25$_{(+0.04)}$ & 13.78& 77.11$_{(\uparrow 0.64)}$ & 62.75$_{(\uparrow 0.84)}$ \\
        \bottomrule
\end{tabular}   
\end{table*}

\begin{table}[htbp]
    \centering
    \scriptsize
     \caption{BIT-based ablation results of components in SSCP. Each sub-module is added sequentially in the order specified in SSCP.}
    \label{BIT_ablation_component}
    \small
    \renewcommand{\arraystretch}{1}
    \setlength{\tabcolsep}{1.8pt}
    \begin{tabular}{cccccccc}
        \toprule
        Method & Dataset & Backbone & MSA & SRGA & CSA & F1(\%) & IoU(\%) \\
        \midrule
        \multirow{9}{*}{SSCP\underline{~}BIT} & \multirow{9}{*}{HSRW-CD} & \multirow{9}{*}{ResNet18} 
        & \xmark & \xmark &  \xmark & 70.03 &  53.88 \\
        
        & & & $\checkmark$ & \xmark & \xmark & 71.61 & 55.78\\
        
        & & & $\checkmark$ & $\checkmark$ & \xmark & 72.19 & 56.48  \\
        
        & & & \xmark & $\checkmark$ & \xmark & 70.35 & 54.27  \\
        
        & & & \xmark & $\checkmark$ & $\checkmark$ & 72.03 & 56.28 \\
        
        & & & \xmark & \xmark & $\checkmark$ & 70.48 & 54.42  \\
        
        & & & $\checkmark$ & \xmark & $\checkmark$ & 71.92 & 56.15  \\

        & & & $\checkmark$ & $\checkmark$ & $\checkmark$ &\textbf{72.49} & \textbf{56.85} \\
        \bottomrule
    \end{tabular}
\end{table}    
\begin{table}[htbp]
    \centering
    \scriptsize
     \caption{BAN-based ablation results of components in SSCP. Each sub-module is added sequentially in the order specified in SSCP.}
    \label{BAN_ablation_component}
    \small
    \renewcommand{\arraystretch}{1}
    \setlength{\tabcolsep}{1.8pt}
    \begin{tabular}{cccccccc}
        \toprule
        Method & Dataset & Backbone & MSA & SRGA & CSA & F1(\%) & IoU(\%) \\
        \midrule
        \multirow{9}{*}{SSCP\underline{~}BAN} & \multirow{9}{*}{HSRW-CD} & \multirow{9}{*}{MiT-B0} 
        & \xmark & \xmark &  \xmark & 74.63 &  59.53 \\
        
        & & & $\checkmark$ & \xmark & \xmark & 75.03 & 60.04\\
        
        & & & $\checkmark$ & $\checkmark$ & \xmark & 75.20 & 60.26  \\
        
        & & & \xmark & $\checkmark$ & \xmark & 74.95 & 59.93  \\
        
        & & & \xmark & $\checkmark$ & $\checkmark$ & 75.59 & 60.75 \\
        
        & & & \xmark & \xmark & $\checkmark$ & 75.28 & 60.36  \\
        
        & & & $\checkmark$ & \xmark & $\checkmark$ & 76.06 & 61.37  \\

        & & & $\checkmark$ & $\checkmark$ & $\checkmark$ &\textbf{76.50} & \textbf{61.94} \\
        \bottomrule
    \end{tabular}
\end{table}    

\begin{figure*}[htbp]
	\centering
	\begin{minipage}{1\linewidth}
		\centering
		\includegraphics[width=0.9\linewidth]{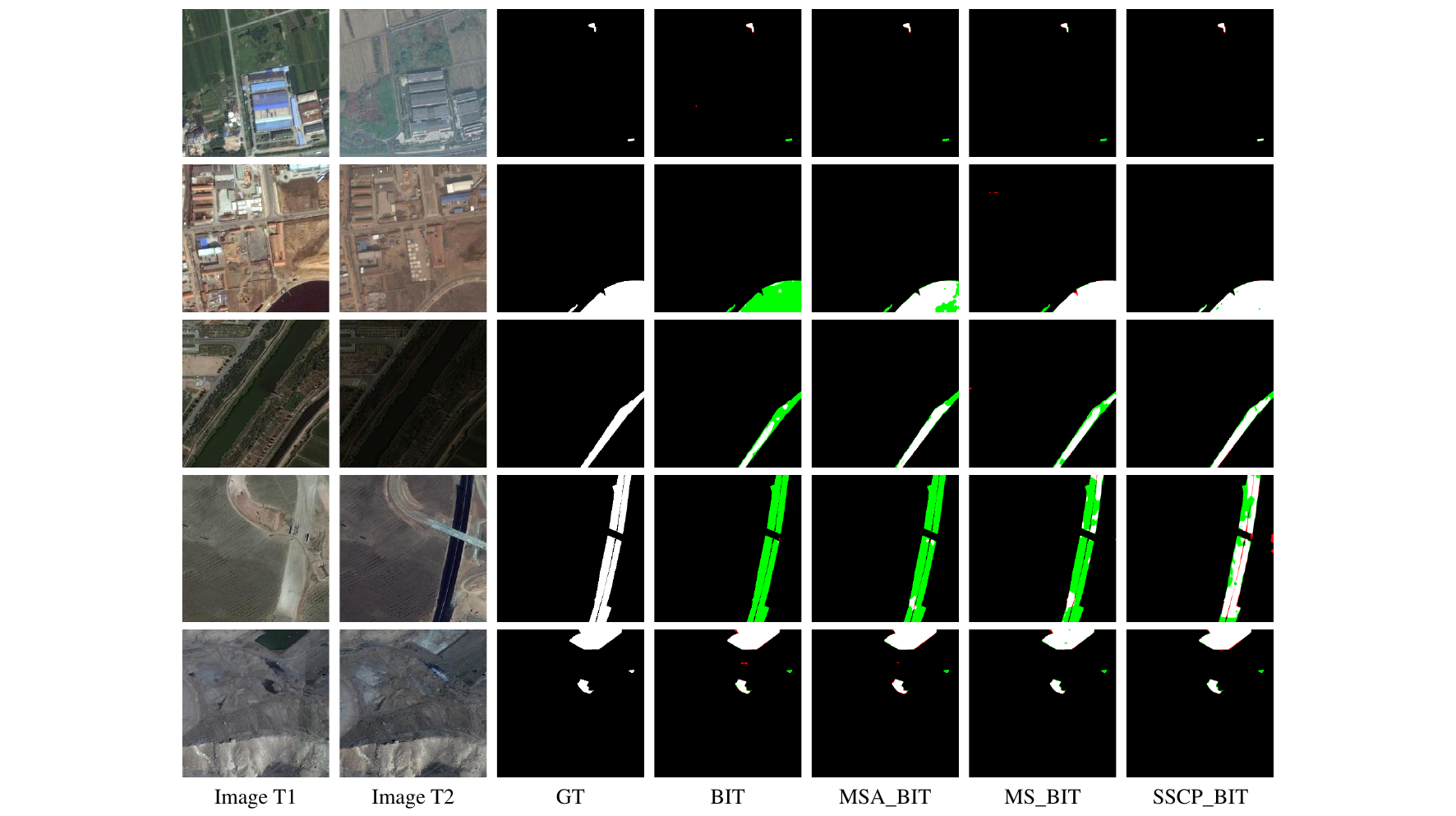}
		\caption{The visualization results of ablation on components in SSCP in SSCP\underline{~}BIT. The MSA\underline{~}BIT represents the BIT fusing MSA. The MS\underline{~}BIT denotes the BIT integrating MSA and SRGA sequentially. The red regions denote the false positive and the green regions are the false negative.}
		\label{BIT_ablation}
	\end{minipage}
 
	\begin{minipage}{1\linewidth}
		\centering
		\includegraphics[width=0.9\linewidth]{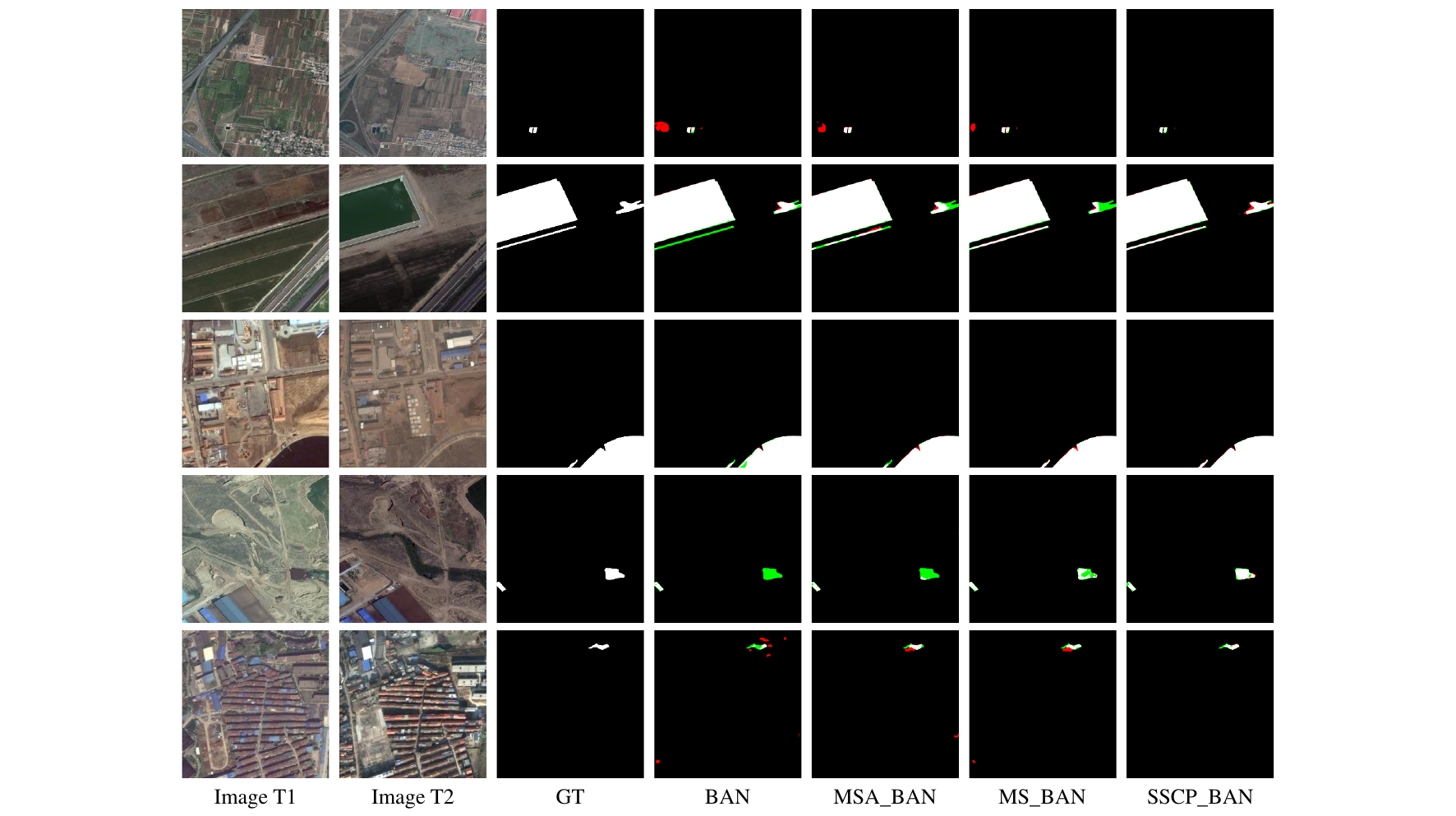}
		\caption{The visualization results of ablation on components in SSCP in SSCP\underline{~}BAN. The MSA\underline{~}BAN denotes the BAN fusing MSA. The MS\underline{~}BAN represents the BAN integrating MSA and SRGA sequentially. The red regions denote the false positive and the green regions are the false negative.}
		\label{BAN_ablation}
	\end{minipage}
\end{figure*}

\textbf{Components}. In this section, we explore the influence of components of SSCP on the HSRW-CD dataset. As illustrated in Table~\ref{BIT_ablation_component} and Table~\ref{BAN_ablation_component}, extensive experiments on SSCP\underline{~}BIT and SSCP\underline{~}BAN are conducted. Through a series of experiments involving different combinations, the individual and synergistic effectiveness of the three components of SSCP are validated. As shown in Table\ref{BIT_ablation_component}, To set the baseline, we initially remove the SSCP. The baseline attains 70.03\% F1 and 53.88\% IoU. Then the MSA is recovered to extract rich spatial semantic information in deep features of the water body, with a 1.58\% F1 and 1.90\% IoU increase relative to the baseline. Subsequently, the SRGA is employed to enhance spatial structure for better continuity of changed information. Gradually,  the individual effects and synergy of modules are studied. Finally, the complete SSCP\underline{~}BIT achieves the best 2.46\% F1 and 2.97\% IoU improvement compared to the baseline model. Similarly, as shown in Table~\ref{BAN_ablation_component}, the whole SSCP\underline{~}BAN with MiT-B0 backbone achieves the best 76.50\% F1 and 61.94\% IoU, which has 1.87\% F1 and 2.41\% IoU improvement over the corresponding baseline. These quantified results also demonstrate the effectiveness of components in SSCP.

Correspondingly, the visual results of ablation on the components in SSCP are presented in Fig.~\ref{BIT_ablation} and Fig.~\ref{BAN_ablation}. It can be observed that the predictions for changed regions tend to be more similar to the ground-truth from left to right. For instance, in the second row in Fig.~\ref{BAN_ablation}, there is an emerging straight canal on the left of Image T2. The MSA obviously enhances the spatial semantics of the canal, while avoiding confusion with high-moisture grasslands exhibiting similar features. Subsequently, the SRGA further improves its spatial continuity. Finally, the CSA leverages the above spatial priors to guide the learning better. As illustrated in Fig.~\ref{BAN-B0_feature_maps_vis}, we present the feature maps passed through three modules sequentially to better understand their effects.It can be observed that the changed regions are located more precisely with the increase of components. Besides, in the first row in Fig.~\ref{BIT_ablation}, Image T2  exhibits unfavorable imaging conditions, which introduces interference into the model's predictions. In Fig.\ref{BAN_ablation}, Image T2 in the last row contains a large number of shadow regions that interfere with the model. Nevertheless, the proposed SSCP attention guides models step by step, enabling their predictions to become more refined and closer to the real changes. The image pairs shown in Fig.~\ref{BIT_ablation} and Fig.~\ref{BAN_ablation} have diverse types and different scales of water body, which further validates the effective adaptation to WBCD of the proposed SSCP. 

\textbf{Key hyperparameters in SSCP}. As Table~\ref{hyperparameter_ablation} lists, we conduct ablation experiments on the important hyperparameters within SSCP. First, single-scale convolutions are attempted. Then, we try multi-scale sub-feature information enhancement. The results demonstrate that MSA with multi-scale feature extraction effectively boosts the spatial semantics of deep features. 
\begin{table}[htbp]
    \centering
    \scriptsize
     \caption{Ablation results of kernel sizes and K (the number of sub-features) in MSA within SSCP.}
    \label{hyperparameter_ablation}
    \scriptsize
    \begin{tabular}{ccccccc}
        \toprule
        Method & Dataset & Backone & Kernel Sizes & K  & F1(\%) & IoU(\%) \\
        \midrule
        \multirow{7}{*}{SSCP\underline{~}BAN} & \multirow{7}{*}{HSRW-CD} & \multirow{7}{*}{MiT-B0}
        & - & - & 74.63 & 59.53 \\
        &&& 3 & 1 & 74.53 & 59.40 \\
        &&& 5 & 1 & 74.81 & 59.76 \\
        &&& 7 & 1 & 75.67 & 60.86 \\
        &&& 3, 7 & 2 & 75.35 & 60.45 \\
        &&& 5, 9 & 2 & 75.66 & 60.85 \\
        &&& 3, 5, 7, 9 & 4 & \textbf{76.50} & \textbf{61.94} \\
        \bottomrule
    \end{tabular}
\end{table}

\textbf{Interdependence between components}. As illustrated in Table~\ref{components_interdependence_analysis}, to explore the interdependence between components in SSCP, we conduct a series of experiments in different orders of module assembly. It can be observed that the model achieves the best F1 and IoU in the order specified in SSCP. Thus, we obtain the optimal order for composing the SSCP attention, aiming to enhance the deep features of water bodies.
\begin{table}[htbp]
    \centering
     \scriptsize
     \caption{The experimental results of interdependence between components in SSCP. Where "$\parallel$" denotes parallel execution, and "Add" represents feature additive fusion (i.e., element-wise addition of feature maps).}
    \label{components_interdependence_analysis}
    \small
    \scriptsize
    \renewcommand{\arraystretch}{1}
    \setlength{\tabcolsep}{1.9pt}
    \begin{tabular}{ccccccc}
        \toprule
        Method &Dataset &  Backbone & Order & F1(\%) & IoU(\%) \\
        \midrule
        \multirow{9}{*}{SSCP\underline{~}BAN} 
        & 
         \multirow{9}{*}{HSRW-CD}  &  \multirow{9}{*}{MiT-B0} & -  & 74.63 &  59.53 \\
        & && SRGA$\rightarrow$ CSA $\rightarrow$ MSA & 76.18 & 61.53\\
        
        & && CSA $\rightarrow$ MSA $\rightarrow$ SRGA& 74.75 & 59.68  \\
        & && CSA $\rightarrow$ SRGA$\rightarrow$ MSA  & 76.19 & 61.54 \\
        & && MSA  $\rightarrow$ CSA $\rightarrow$ SRGA & 75.01 & 60.01  \\
        & && SRGA $\rightarrow$ MSA $\rightarrow$ CSA & 75.13 & 60.17  \\
        & & & (MSA $\parallel$ SRGA)$\rightarrow$ Add $\rightarrow$ CSA & 75.20 & 60.26  \\
        && & SSCP &\textbf{76.50} & \textbf{61.94} \\
        \bottomrule
    \end{tabular}
\end{table}

\textbf{Performance comparison with existing attention mechanisms}. To further validate the advantages of the proposed SSCP, we compare the performance of the SSCP with existing efficient attention mechanisms  in Table~\ref{attention_comparison}. The proposed SSCP attention exhibits the best F1 and IoU, while maintaining parameters and FLOPs comparable to the other attention mechanisms. It achieves the optimal trade-off between performance and efficiency, which demonstrates the superiority of the SSCP for leveraging water body deep features to improve WBCD research.
\begin{table*}[ht]
    \centering
    \scriptsize
     \caption{{Comparison results of the proposed SSCP with other attention mechanisms to refine water body deep features.}}
    \label{attention_comparison}
    \small
    \scriptsize
    \renewcommand{\arraystretch}{1}
    \setlength{\tabcolsep}{2pt}
    \begin{tabular}{ccccccccc}
        \toprule
        Method & Dataset &  Backbone & Attention & F1(\%) & IoU(\%) & Params(M) &  FLOPs (G) \\
        \midrule
        \multirow{9}{*}{BAN~\citep{li2024ban}} &\multirow{9}{*}{HSRW-CD}  &  \multirow{9}{*}{MiT-B0} & - & 74.63 &  59.53 & 232.07 & 133.97 \\
        & && ECA~\citep{wang2020eca} & 74.09 & 58.84 & 232.07 & 133.97  \\
        & & & SPP~\citep{he2015spatial} & 74.91 & 59.89 & 232.24 & 134.05\\
        &&& CAA~\citep{cai2024poly}& 75.35 & 60.45 & 232.21 & 134.04 \\
        &&& ELA~\citep{xu2024ela}& 75.20 & 60.25 & 232.08 & 133.97\\
        &&& CA~\citep{hou2021coordinate} & 75.26 & 60.33 & 232.12 & 133.97 \\
        &&& CASAtt~\citep{zhang2024cas} & 75.72 & 60.92 & 232.41 & 134.07 \\
        &&& SSCP &\textbf{76.50} & \textbf{61.94} & 232.11 & 133.99\\
        \bottomrule
    \end{tabular}
\end{table*}    

\begin{figure*}[htbp]
\centering
\includegraphics[width=1\textwidth]{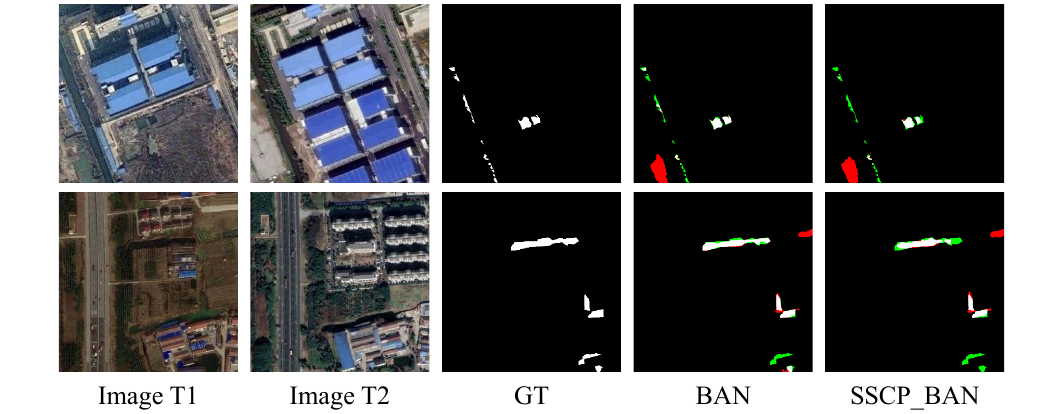}
\caption{The visualization results of failure cases on the test set of HSRW-CD based on BAN with MiT-B2 backbone. The red regions denote the false positive, while the green regions are the false negative.}
\label{failure_cases}
\end{figure*}
\subsection{Failure case analysis}
As shown in Fig.~\ref{failure_cases}, for high-moisture vegetation in images that are highly similar to water bodies, the proposed SSCP attention fails to capture subtle features, thereby leading to false positives. For instance,  in the lower left corner of Image T1 in the first row, the high-moisture farmland exhibits smooth textural features similar to certain water bodies in the visible band. However, corresponding deep features suffer from limited spatial detail information. As a result, the SSCP attention which relies on these deep features for discriminative weighting, falls short in prioritizing the subtle distinguishing cues. To address this issue, future research will further explore a multi-scale feature refinement network to obtain differentiating  features.

\section{Conclusion}
This paper first  put forward the HSRW-CD dataset, which addresses the scarcity of high spatial resolution datasets for remote sensing WBCD. In addition, a Spatial Semantics and Continuity Perception (SSCP) attention module is specially designed to refine deep features of the water body for WBCD. Specifically, the proposed SSCP consists of the Multi-Semantic Spatial Attention (MSA), Structural Relation-Aware Global Attention (SRGA), and Channel-wise Self-Attention (CSA) to improve the performance of WBCD. The MSA divides the deep features to extract varying spatial semantic information, effectively learning the semantics of different water body types. Subsequently, the SRGA further models the spatial structural relation to provide the CSA with spatial structure priors. Finally, the CSA fully utilizes the information from the MSA and SRGA to reweight channels. The SSCP suppresses the semantic expression of irrelevant features while improving the spatial structural continuity of changed regions. Two WBCD-specific benchmark datasets are used to conduct comprehensive experiments, verifying the effectiveness and generalization of the proposed SSCP.

\textbf{Limitations and Future Work}. However, in a water body bi-temporal image pair, the proportion of spatial semantic  and structural information in deep features needs to be further explored. Besides, the proposed SSCP attention in this paper is merely a plug-and-play module, which leverages the rich information in deep features. In the future, we will design a complete change detection network specifically for WBCD based on the characteristics of water bodies and failure cases. 





\bibliographystyle{elsarticle-num}
\bibliography{references}

\end{document}